\documentclass[]{article} 
\usepackage{siunitx,graphicx,subfigure,overpic,amsmath,amssymb,bm}
\usepackage{dcolumn}  	
\usepackage{authblk}
\usepackage{verbatim}   	
\usepackage{color}
\newcommand{\op}[1]{\ensuremath{\mathcal{#1}}}

\begin{document}

\title{Analysis of the noise in back-projection light field acquisition and its optimization }

\author[1,2,*]{Ni Chen}
\author[2]{Zhenbo Ren}
\author[1]{Dayan Li}
\author[2]{Edmund Y. Lam}
\author[1]{Guohai Situ}
\affil[1]{Shanghai Institute of Optics and Fine Mechanics, Chinese Academy of Sciences, Shanghai 201800, China}
\affil[2]{Department of Electrical and Electronic Engineering, The University of Hong Kong, Pokfulam, Hong Kong.}

\affil[*]{Corresponding authors: nichen@siom.ac.cn}
   
\date{\today}
\maketitle

\begin{abstract}
Light field reconstruction from images captured by focal plane sweeping can achieve high lateral resolution comparable to the modern camera sensor. This is impossible for the conventional micro-lenslet based light field capture systems. However, the severe defocus noise and the low depth resolution limit its applications. In this paper, we analyze the defocus noise and the depth resolution in the focal plane sweeping based light field reconstruction technique, and propose a method to reduce the defocus noise and improve the depth resolution. Both numerical and experimental results verify the proposed method.
\end{abstract}

\section{Introduction}
\begin{figure}[t]
  \centering
  \begin{minipage}{.26\linewidth}
  \includegraphics[width=1\columnwidth]{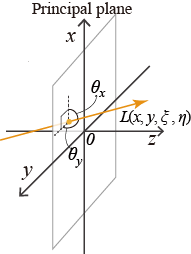}
  \centerline{(a)}
  \end{minipage}
  \begin{minipage}{.2\linewidth}
  \includegraphics[width=1\columnwidth]{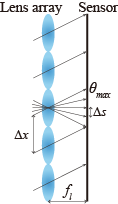}
  \centerline{(b)}
  \end{minipage}
\caption{\label{fig_lf} (a) Light field definition, and (b) micro-lens array based light field capture.}
\end{figure}

A single view image of a three-dimensional~(3D) scene corresponds to the projection of a collection of light rays coming from it, and a light ray with its propagation direction is called a light field~\cite{Gershun_1936,Lam_2015_JOSA}. Unlike conventional photography, which records only the intensity distribution of the light rays, a light field camera~\cite{Ng_2005_ACM} records both the intensity and direction of the light rays~\cite{Levoy_2009_JM}, enabling view reconstruction of the 3D properties of a scene~\cite{Ng_2006}. In Fig.~\ref{fig_lf}(a), we describe a five-dimensional~(5D) light field function~\cite{Levoy_2009_JM}. The principal plane of the light field is perpendicular to the optical axis. Many light rays with different directions go through each position on the principal plane, and every ray can be fully described by a 5D function $L(x,y,\xi, \eta, z)$, where $(x,y)$ is the lateral position of the light field at the principal plane located at depth $z$, $(\xi=\theta_x, \eta= \theta_y)$ are the projection angles between the light ray and the normal of the principal plane. Usually, a four-dimensional~(4D) function without the $z$ coordinate is enough to represent a light field~\cite{Levoy_1996_LFR}, because the light field can be assumed to propagate along the optical axis. In this paper, we use the 4D light field function. Since light field records the 3D information of a scene, and because of the storage of the direction of each light ray, it thus can be used for many applications, such as refocusing~\cite{Ng_2005_ACM}, and auto-stereoscopic 3D display, i.e. integral imaging~\cite{Hong_2011_AO,Park_2014_JID}. Besides, it can also be used to synthesize a hologram to eliminate the coherence requirement in hologram recording systems~\cite{Park_2009_OE,Chen_2014_IS,Chen_2016_AO}. 

In most light field capture techniques, a camera with a micro-lens array in front of its sensor~\cite{Levoy_2009_JM,Ng_2005_ACM,Ng_2006} is used, where every micro-lens captures angular distribution of the light rays at its principal point, as Fig.~\ref{fig_lf}(b) shows. The number of light rays that can be recorded depends on the lens pitch $\Delta_x$ and the pixel pitch $\Delta_s$ of the camera sensor. The maximum angle $\theta_{max}$ of the light rays that can be collected depends on the specification of the micro-lens, i.e., the focal length $f_l$ and the lens pitch $\Delta x$. The spatial sampling interval of the object is the same as the pitch of the lens array. This lens array based method enables direct capture of the light field at a single shooting. However, the spatial resolution and angular resolution of the captured light field mutually restrict each other, therefore the achieved spatial resolution is much lower than that of the image sensor~\cite{Hong_2011_AO,Park_2014_JID}. Although several methods have been proposed to enhance the spatial resolution, they usually require to solve a computationally heavy inverse problem, sometimes with prior knowledge of the object scene~\cite{Chen_2010_OE,Chen_2011_OE}.

Recently, it has been reported that the light field can also be obtained from focal plane sweeping captured images with a conventional digital camera ~\cite{Orth_2013_OL,Park_2014_OE}. These techniques can capture a higher resolution light field. Examples are the light field moment imaging~(LFMI)~\cite{Orth_2013_OL,Liu_2015_OE} and the light field reconstruction with back projection~(LFBP) approach~\cite{Park_2014_OE}. In these cases, the light field is calculated from several photographic images captured at different focus depths, the images are not segmented by the sub lens of the lens array, hence they can reach a higher angular and spatial image resolution comparable to that of a conventional camera sensor. Note that the angular sampling of the light field calculated from the photographic images depends on the numerical aperture~(NA) and the pixel pitch of the camera sensor, rather than the number of images captured along the optical axis. As these methods do not require any special equipment like lens array, it is easy to be implemented. Capturing at a fixed camera location reduces the burden of the calibration greatly. Finally, the estimated light ray field can have high spatial resolution comparable to that of the image sensor itself.

Although the LFBP can reconstruct an exact light field with high angular resolution, severe defocus noise exists in the reconstruction~\cite{Park_2014_OE,Chen_2016_AO}. In this paper, we analyze the noise in the reconstructed light field. Besides, we propose a method to suppress the noise. Numerical and experimental data are also presented to verify the proposed method.


\section{Defocus noise analysis of the focal plane sweeping based light field reconstruction}

\subsection{Light field reconstruction from focal plane sweeping captured images}
\begin{figure}[t]
  \centering
  \includegraphics[width=.6\columnwidth]{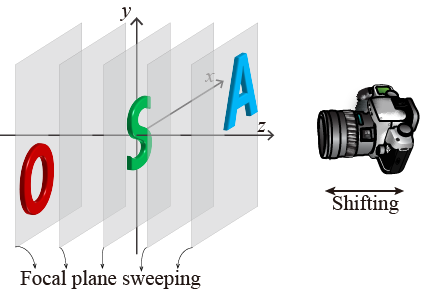}
\caption{\label{fig_lfbp} Scheme for capturing focal plane sweeping images.}
\end{figure}

In the LFBP technique, a series of images along the optical axis are captured while the camera's focal plane is shifted, as Fig.~\ref{fig_lfbp} shows. The focal plane shifting can be achieved by turning the focus ring of the camera. The image $\op{I}(x, y, z_m)$ is captured while the focal plane is located at $z=z_m$. The total number of captured images denoted as $M$. Generally, the focal plane sweeping range should cover the depth range of the 3D scene. With these captured images, the light field with the principal plane located at $z=0$ is calculated by using the back-projection algorithm~\cite{Park_2014_OE}
\begin{align}
L(x, y, \xi, \eta) =  \sum\limits_{m=1}^{M}{I(x+z_m \xi, y+z_m \eta, z_m)}.
\label{eq_lfbp}
\end{align}
Here we omit the magnification factor of the images. This is because the captured images can be aligned and resized easily with post digital processing. The light field reconstructed with this approach has a severe noise problem~\cite{Park_2014_OE,Chen_2016_AO}. In order to eliminate the noise, we should study its origin. The mathematical analysis is shown in the following section.

\subsection{Analysis of the defocus noise in the LFBP reconstructed light field}
\begin{figure}[htb]
   \centering
   \includegraphics[width=.35\columnwidth]{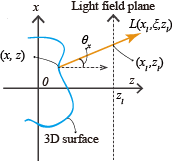}
\caption{\label{fig_obj_lf} Relation between 3D object and its light field.}
\end{figure}
We start from considering a 3D object with its center located at the origin of the Cartesian coordinates. Its surface function can be approximately represented as a stack of object slices, i.e., $\op{O}(x,y,z)\approx\sum_{n=1}^{N}\op{O}(x,y,z_n)$, where $\op{O}(x,y,z_n)$ is the object slice located at a depth of $z=z_n$, and $N$ is the total number of the object slices. It should be noticed that $z_n$ here is different from $z_m$ in Eq.(\ref{eq_lfbp}), while $z_n$ is the discrete sampled depth of the 3D object, and $z_m$ is the focal plane of the camera used to capture the 3D object. Since the energy traveling along a ray is considered as a constant, the light field is the integral of the object projections, as Fig.~\ref{fig_obj_lf} shows. The light field with the principal plane located at the center of the object thus can be represented as~\cite{Chen_2016_AO}
\begin{align}
L(x, y, \xi, \eta) = \sum_{n=1}^{N}{\op{O} (x+z_n\xi, y+z_n\eta, z_n)}.
\label{eq_lf}
\end{align}
When we capture an image of a 3D scene, suppose the camera focal plane is located at $z=z_m$, the captured images should be the convolution of the clear images of the object and the camera's point spread function. The captured image with the camera focal plane at $z=z_m$ is marked as $\op{I}(x,y, z_m)$, and its math representation is

\begin{align}
\op{I}(x,y, z_m)
 = &\sum\limits_{n=1}^{N} {\op{O}(x,y, z_n) \otimes h(x,y, z_m-z_n)} \nonumber \\
 = & \op{O}(x, y, z_m) + \sum\limits_{n=1}^{N, z_n \neq z_m} {\op{O}(x,y, z_n) \otimes h(x,y, z_m-z_n)},
\label{eq_I}
\end{align}
where $h(x,y,z_m-z_n)$ is the Point Spread Function~(PSF) of the camera, and $\otimes$ is the two dimensional convolution operator. The first term in the second line of Eq.~(\ref{eq_I}) is the clear image of the object slice at the focal plane, the second and third terms are the blurred image contributed by the object slices which are out of focus. Substituting Eq.~(\ref{eq_I}) to Eq.~(\ref{eq_lfbp}) we obtain the equation of the LFBP reconstructed light field $L^\prime(x, y, \xi, \eta)$ as
\begin{align}
& L^\prime(x, y, \xi, \eta) \nonumber\\
= &\sum\limits_{m=1}^{M}{\op{O}(x+z_m \xi, y+z_m \eta, z_m)} \nonumber \\
  &+ \sum\limits_{m=1}^{M}\left\{\sum\limits_{n=1}^{N,z_n\neq z_m}{\op{O}(x+z_m \xi, y+z_m \eta, z_n)} \otimes{h(x+z_m \xi, y+z_m \eta, z_m-z_n)}\right\}.
\label{eq_lf_prime}
\end{align}
\noindent In Eq.~(\ref{eq_lf_prime}), suppose the number of the captured images is comparable to the number of the object slice, i.e. $M\approx N$. In this case, the first term in Eq.~(\ref{eq_lf_prime}) equals to Eq.~(\ref{eq_lf}), which corresponds to the exact light field of the 3D object. The second term is the defocus noise. Obviously, it is the accumulation of the defocus noise induced by the images of the object slices which are out of focus. From this equation, we can see that there are two main parameters affecting the noise: the number of the depth images and the PSF of the camera. The PSF is related to the f-number of the camera, in other words, the Numerical Aperture~(NA). In order to view how the parameters affect the defocus noise, we calculate the noise with respect to the two parameters sequentially.    

\begin{figure}[!htb]
  \centering
  \begin{minipage}{.45\linewidth}
  \includegraphics[width=1\columnwidth]{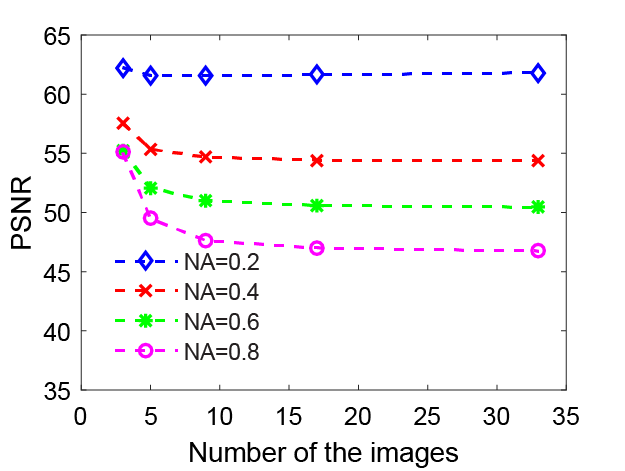}
  \centerline{(a)}
  \end{minipage}
  \begin{minipage}{.45\linewidth}
  \includegraphics[width=1\columnwidth]{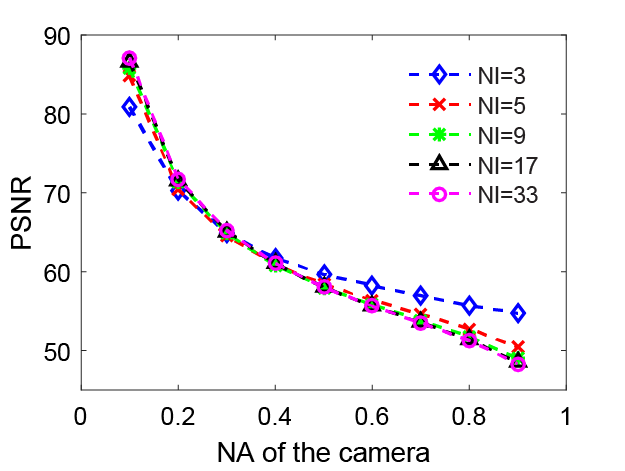}
  \centerline{(b)}
  \end{minipage}
\caption{Noise in the reconstructed light field with respect to (a) the NA of the camera and (b) the number of captured images.}
\label{fig_sim_noise}
\end{figure}

A 3D object with three planes while each of them has a pixel number of $256 \times 256$ is used in the calculation. The depth interval between two adjacent object slices is $\SI{20}{\milli\meter}$, and the center of the object is located at the origin of the chosen Cartesian coordinate. The PSF of the camera is supposed to be a Gaussian function.

In Fig.~\ref{fig_sim_noise}(a), the Peak Signal to Noise Ratio~(PSNR) of the reconstructed light field is plotted in respect with the number of the captured images. In the numerical captures, the depth range that covers the whole object is constant, only the distance between two photos was modified to capture different number of images. Several results were obtained according to different camera NAs. 
The figure shows the PSNR is decreasing as the increasing number of captured images, i.e., the noise become more and more severe with the increasing number of the captured images. This can also be observed from Eq.~(\ref{eq_lf_prime}), more images lead to further noise accumulation in the second term. 
In Fig.~\ref{fig_sim_noise}(b), we plot the PSNR of the reconstructed light field with respect to the NA of the camera. Several groups of images with various numbers~(NI in the figure) of photos were numerically captured and used for the light field reconstruction. It is obvious that the PSNR is decreasing with the increasing NA of the camera no matter how many photos were used to reconstruct the light field. We can also say the noise is increasing with the increasing NA of the camera. Generally, the PSF of a camera is cone-shaped and symmetrical about the focal plane. A smaller camera NA produces a slimmer cone-shaped PSF, as well as less noise accumulation in the second term of Eq.~(\ref{eq_lf_prime}). 

From Fig.~\ref{fig_sim_noise}, it is clear that in the LFBP technique, smaller camera NA and fewer images produce higher quality reconstructed light field. However, in order to maintain the depth resolution of the reconstructed light field, the number of the captured images should be large enough. This mutual constraint property limits the conventional LFBP technique can be only applied to objects of a few loosen slices. In the following section, we show our improved LFBP technique, which solves this problem.

\section{Optimization of the focal plane sweeping based light field reconstruction}

\subsection{Principle of the optimization method}
In the previous section, we show that the noise in the reconstructed light field comes from the accumulation of defocus noise of the captured images. In order to eliminate the defocus noise, we should detect the defocus noise first. Since we capture the images along the optical axis with the focal plane sweeping approach, the sharp image area in one image will be blurred in all the other images. The amount of changes between two adjacent images reflect the sharpness degree of an image, and the highest sharpness degree indicates the clearest image location. Therefore, detecting the maximum changing of each pixels along the optical axis can help us find the focus and out of focus part in the captures images. In photography deblurring, Laplace operator is an efficient approach, which detects the gradient changes of an image. Here we apply this technique to achieve our aim. The detected focus pixels in each captured images are then combined as the new images to calculate the light field. Since the redundant defocus noise is omitted before the calculation, the quality of the reconstructed light field is improved.

During the preprocessing, the captured images are treated as an image stack $\bm{\op{I}_s(x,y,z)}$
\begin{align}
 \bm{\op{I}_s(x,y,z)}=\left[
 \begin{matrix}
   \op{I}(x,y,z_1) \\
   \vdots \\
   \op{I}(x,y,z_M)
  \end{matrix}
  \right].
\end{align}
Laplace operator is used to detect the edge of the image. However, it is sensitive to discrete points and noise. Therefore we filter the images with Gaussian filter to reduce the noise, this can increase the robust of the Laplace operator.
\begin{align}
\bm{E}_s(x,y,z) = \nabla^2\left\{\bm{\op{I}_s(x,y,z)} \otimes G(x,y)\right\},
\label{eq_I_stack}
\end{align}
where $\displaystyle{G(x,y) = \frac{1}{2 \pi \sigma^2}\exp\left[-\frac{1}{2\pi \sigma^2}(x^2+y^2) \right]}$ is the Gaussian filter to smooth the images. The Gaussian kernel size depends on what is desired, large kernel size detects large scale edges, small kernel size detects fine features. As the derivative is used to measure changes, derivative having maximum magnitude is the information we are looking for
\begin{align}
z_{max}(x,y) = \arg \max_{z}\left\{\bm{E}_s(x,y,z)\right\},
\label{eq_stack_edge}
\end{align}
where $z_{max}(x,y)$ is the position where $\bm{E_s(x,y,z)}$ has the maximum value. The image stack after the pre-processing thus is
\begin{align}
\bm{\op{I}_s}^\prime(x,y,z^\prime) = \bm{\op{I}_s}(x,y,z_{max}(x,y)).
\label{eq_I_prime}
\end{align}
The new image stack $\bm{\op{I}_s}^\prime(x,y,z^\prime)$ is used to synthesize the light field with the same method as in the conventional method.

\section{Simulation verification}
\begin{figure}[ht]
  \centering
  \begin{minipage}{.32\linewidth}
  \includegraphics[width=1\columnwidth]{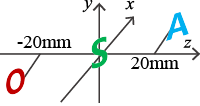}
  \centerline{(a)}
  \end{minipage}
  \hspace{5pt}
  \begin{minipage}{.28\linewidth}
  \vspace{15pt}
  \includegraphics[width=1\columnwidth]{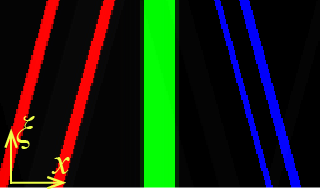}
  \centerline{(b)}
  \end{minipage}
\caption{\label{fig_sim_obj} (a) Object used in the simulation, and (b) an EPI image of the corresponding light field.}
\end{figure}

A 3D object scene with three plane images are used to test our method. The lateral size of each plane is $\SI{128}{\milli\meter} \times \SI{128}{\milli\meter}$, and the distance between two plane images is $\SI{20}{\milli\meter}$. Since the centre of the object is located at the origin of the coordinates, the depth of the three planes are $\SI{-20}{\milli\meter}$, $\SI{0}{\milli\meter}$ and $\SI{20}{\milli\meter}$ respectively. In the simulation, the exact light field of the object scene can be obtained by projecting all the pixels to the principal plane with Eq.~(\ref{eq_lf}), this can be used as the ground truth for comparison. Fig.~\ref{fig_sim_obj}(b) is one of the epipolar-plane image (EPI) image~($L(x,0,\xi,0)$) profile of the light field.

\begin{figure}[htb]
  \centering
  \vspace{1pt}
  \begin{overpic}[width=.2\linewidth]{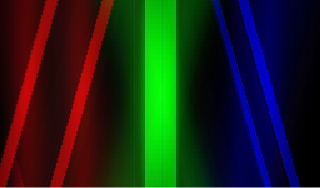}
    \put(-15,5){\rotatebox[origin=tl, units=360]{90}{(a)}}
  \end{overpic}
  \begin{overpic}[width=.2\linewidth]{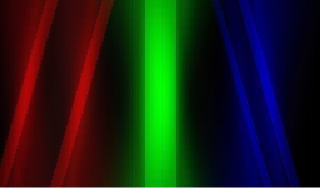}
  \end{overpic}
  \begin{overpic}[width=.2\linewidth]{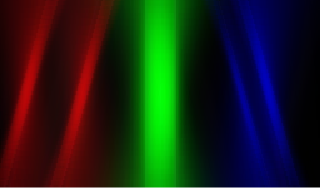}
  \end{overpic}
  \begin{overpic}[width=.2\linewidth]{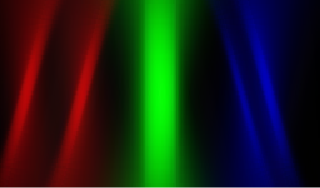}
  \end{overpic}

  \vspace{3pt}
  \begin{overpic}[width=.2\linewidth]{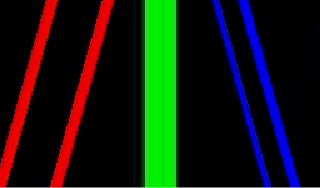}
    \put(-15,5){\rotatebox[origin=tl, units=360]{90}{(b)}}
    \put(40,-15){\color{black}{N=3}}
  \end{overpic}
  \begin{overpic}[width=.2\linewidth]{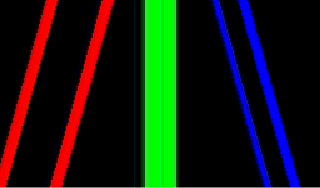}
    \put(40,-15){\color{black}{N=5}}
  \end{overpic}
  \begin{overpic}[width=.2\linewidth]{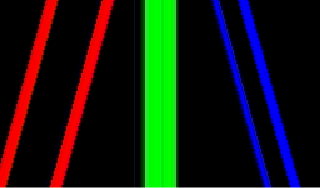}
    \put(40,-15){\color{black}{N=9}}
  \end{overpic}
  \begin{overpic}[width=.2\linewidth]{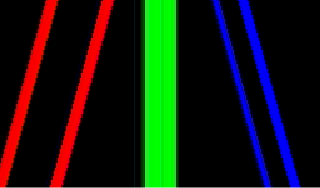}
    \put(40,-15){\color{black}{N=17}}
  \end{overpic}

  \vspace{5pt}
\caption{The reconstructed EPI images of the light field from various number of photos with (a) the conventional method and (b) the proposed method respectively.}
\label{fig_sim_lf_NI}
\end{figure}

\begin{figure}[htb]
  \centering

  \vspace{1pt}
  \begin{overpic}[width=.2\linewidth]{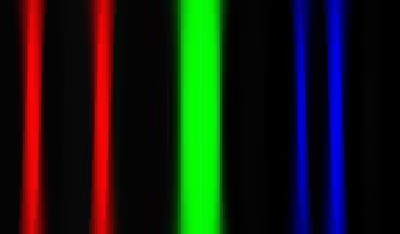}
    \put(-15,5){\rotatebox[origin=tl, units=360]{90}{(a)}}
  \end{overpic}
  \begin{overpic}[width=.2\linewidth]{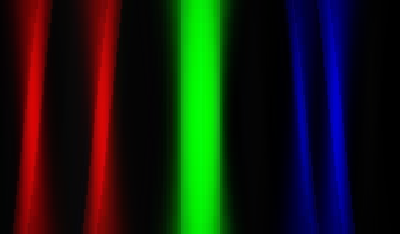}
  \end{overpic}
  \begin{overpic}[width=.2\linewidth]{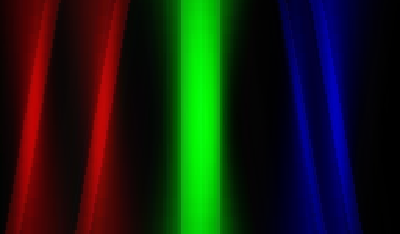}
  \end{overpic}
  \begin{overpic}[width=.2\linewidth]{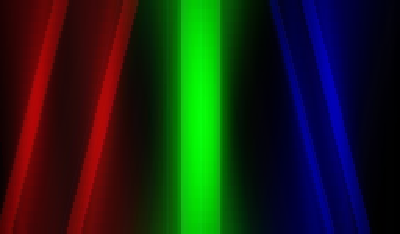}
  \end{overpic}

  \vspace{3pt}
  \begin{overpic}[width=.2\linewidth]{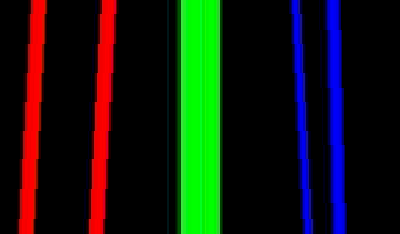}
    \put(-15,5){\rotatebox[origin=tl, units=360]{90}{(b)}}
    \put(25,-15){\color{black}{NA=0.2}}
  \end{overpic}
  \begin{overpic}[width=.2\linewidth]{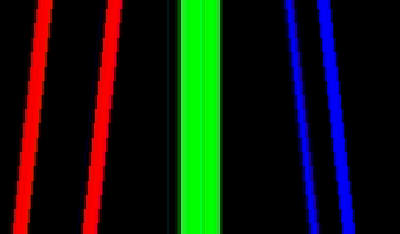}
    \put(25,-15){\color{black}{NA=0.4}}
  \end{overpic}
  \begin{overpic}[width=.2\linewidth]{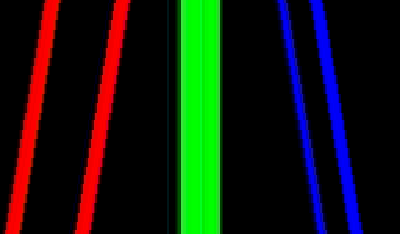}
    \put(25,-15){\color{black}{NA=0.6}}
  \end{overpic}
  \begin{overpic}[width=.2\linewidth]{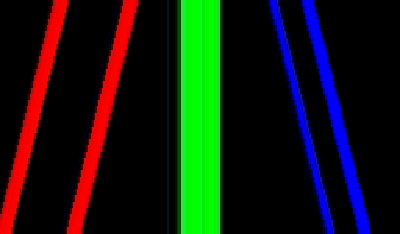}
    \put(25,-15){\color{black}{NA=0.8}}
  \end{overpic}

  \vspace{5pt}
\caption{The reconstructed EPI images of the light field calculated from 5 photos captured under various camera NAs with (a) the conventional method and (b) the proposed method respectively.}
\label{fig_sim_lf_NA}
\end{figure}

The captured photographic images are calculated with Eq.~(\ref{eq_I}). The light field is reconstructed with the conventional and the proposed methods respectively. In Fig.~\ref{fig_sim_lf_NI}(a) and (b) are the conventional and proposed light field reconstructions with different number of captured images. Fig.~\ref{fig_sim_lf_NA}(a) and (b) are the conventional and proposed light field reconstructions with different camera NAs. As explained in Section 2, the noise in the conventional reconstruction get worse with increasing number of the captured images and camera NAs. Conversely, the number of the captured images and camera NAs do not affect the reconstruction of the proposed method.

\begin{figure}[!ht]
  \centering
  \begin{overpic}[width=.18\linewidth]{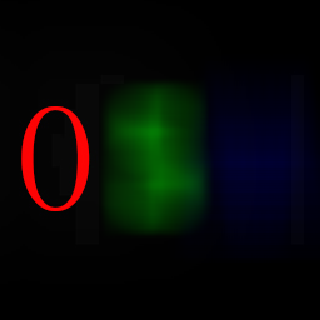}
    \put(-15,5){\rotatebox[origin=tl, units=360]{90}{$z=\SI{-20}{\milli\meter}$}}
  \end{overpic}
  \begin{overpic}[width=.18\linewidth]{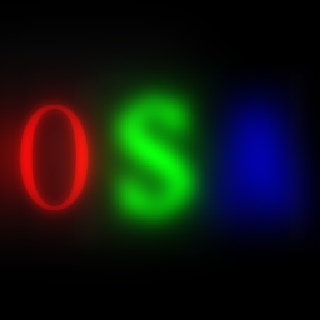}
  \end{overpic}
  \begin{overpic}[width=.18\linewidth]{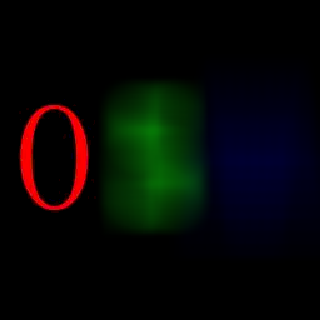}
  \end{overpic}

  \vspace{1pt}
  \begin{overpic}[width=.18\linewidth]{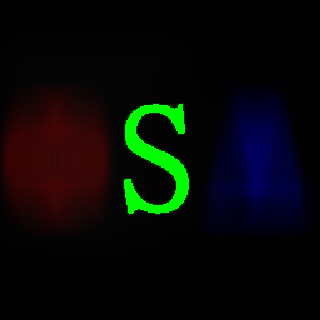}
    \put(-15,15){\rotatebox[origin=tl, units=360]{90}{$z=\SI{0}{\milli\meter}$}}
  \end{overpic}
  \begin{overpic}[width=.18\linewidth]{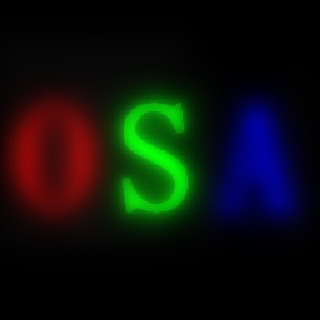}
  \end{overpic}
  \begin{overpic}[width=.18\linewidth]{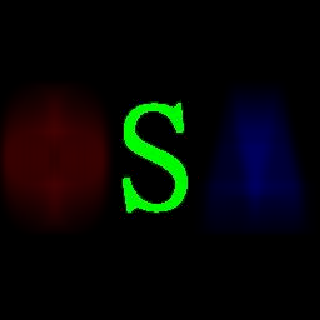}
  \end{overpic}

  \vspace{1pt}
  \begin{overpic}[width=.18\linewidth]{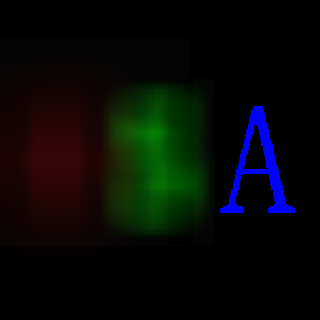}
    \put(-15,10){\rotatebox[origin=tl, units=360]{90}{$z=\SI{20}{\milli\meter}$}}
    \put(40,-15){\color{black}{(a)}}
  \end{overpic}
  \begin{overpic}[width=.18\linewidth]{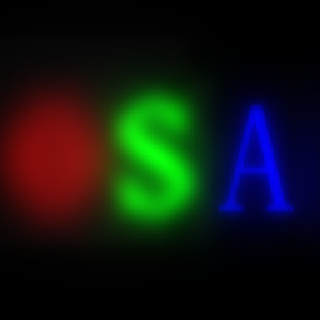}
    \put(40,-15){\color{black}{(b)}}
  \end{overpic}
  \begin{overpic}[width=.18\linewidth]{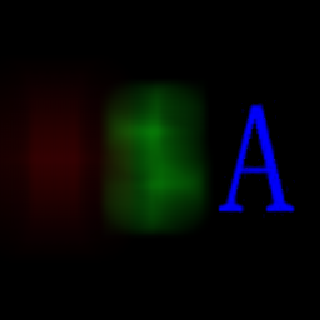}
    \put(40,-15){\color{black}{(c)}}
  \end{overpic}

  \vspace{5pt}
\caption{Refocused images at the three planes of the original object slices from (a) the exact light field and the light field calculated with (b) the conventional method and (c) the proposed method.}
\label{fig_sim_refocus}
\end{figure}

The visualization in Fig.~\ref{fig_sim_lf_NI} and Fig.~\ref{fig_sim_lf_NA} are only a slice of the 4D light field, therefore we also compare the refocused images from the whole light field. A light field with dimensions of $256\times256\times50\times50$ calculated from 5 photographic images are used to perform the refocusing. Fig.~\ref{fig_sim_refocus} shows the refocused images located at the three depths of the original object planes, while Fig.~\ref{fig_sim_refocus}(a), (b) and (c) are corresponding to the exact, the conventional and the proposed method respectively. It is obvious that the focused images with the conventional calculated light field are so blurred, we even can not identify which one is in focus from Fig.~\ref{fig_sim_refocus}(b). On the contrary, the refocused images with the proposed method in Fig.~\ref{fig_sim_refocus}(c) reach clear focus images comparable to the one in Fig.~\ref{fig_sim_refocus}(a). 

\subsection{Experimental verification}
\begin{figure}[!htb]
   \centering
   \includegraphics[width=.4\columnwidth]{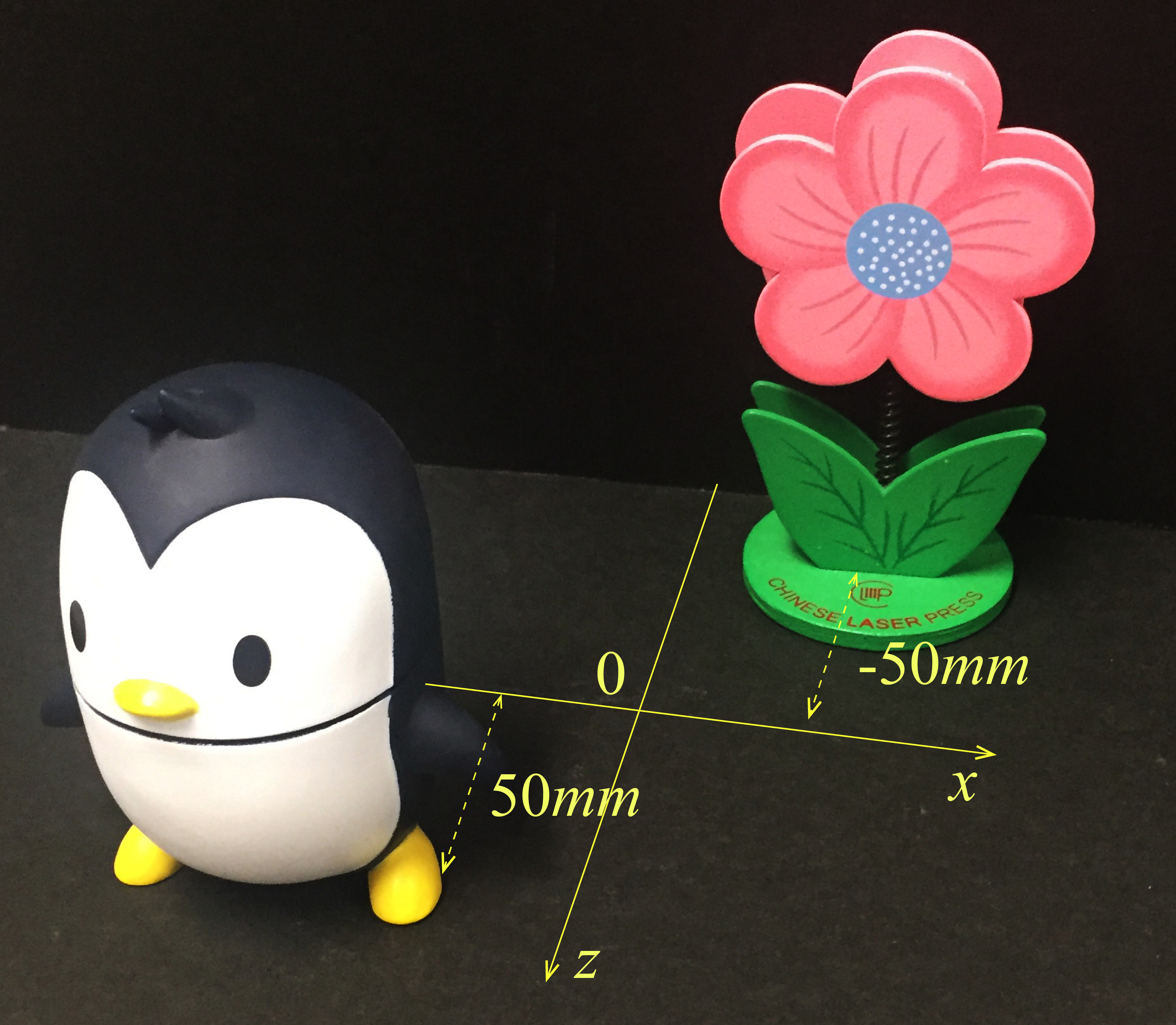}
\caption{\label{fig_exp_obj} Image of the target 3D scene in the experiment.}
\end{figure}
Our proposed method has also been verified with the real captured images. Fig.~\ref{fig_exp_obj} is the image of the objects used in the experiment. A penguin doll and a flower were separated with a depth distance of \SI{100}{\milli\meter}. A Canon EOS 1100D camera was used to take the photos. The NA of the camera is 0.4 and the sensor pixel pitch is \SI{3.1}{\micro\meter}. Three group of images were taken with 3, 5 and 11 photos in each group. All the images were cropped to a resolution of $500\times 400$ pixels for reducing the computational load.
\begin{figure}[htb]
  \centering

  \vspace{1pt}
   \begin{overpic}[width=.3\linewidth]{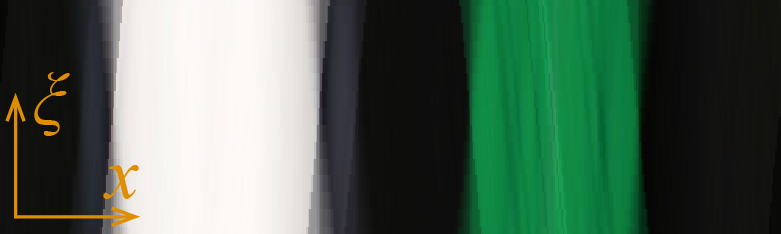}
    \put(-15,0){\rotatebox[origin=tl, units=360]{90}{(a)}}
  \end{overpic}
  \begin{overpic}[width=.3\linewidth]{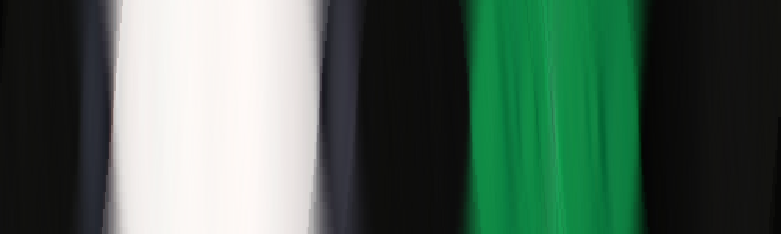}
  \end{overpic}
  \begin{overpic}[width=.3\linewidth]{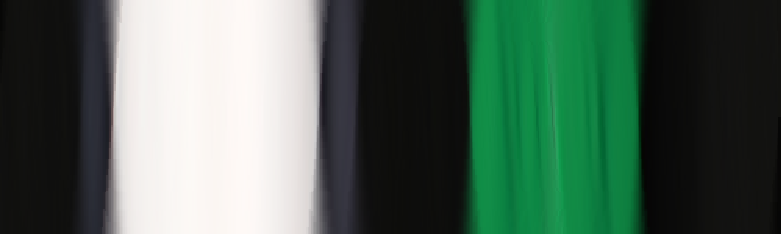}
  \end{overpic}

  \vspace{2pt}
  \begin{overpic}[width=.3\linewidth]{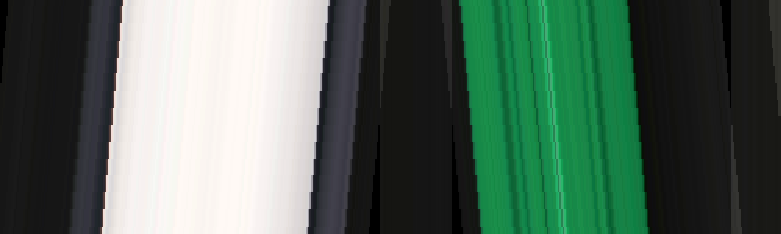}
    \put(-15,0){\rotatebox[origin=tl, units=360]{90}{(b)}}
  \end{overpic}
  \begin{overpic}[width=.3\linewidth]{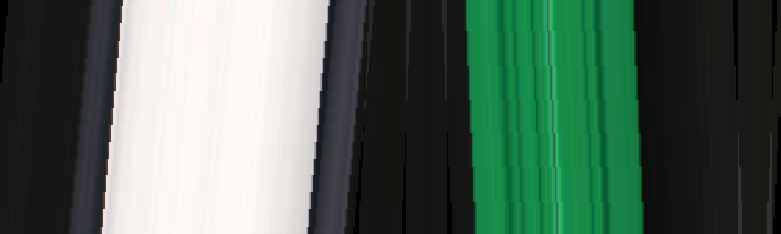}
  \end{overpic}
  \begin{overpic}[width=.3\linewidth]{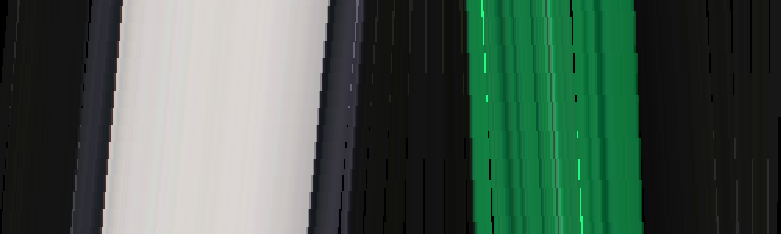}
  \end{overpic}

  \vspace{5pt}
  \begin{overpic}[width=.3\linewidth]{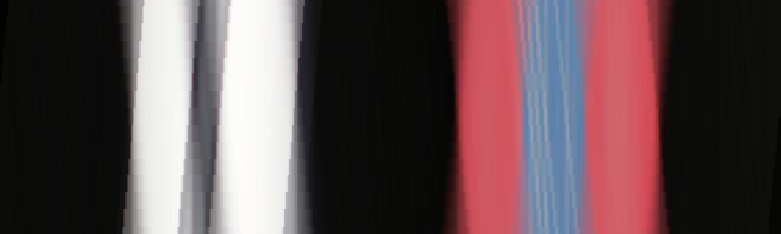}
    \put(-15,0){\rotatebox[origin=tl, units=360]{90}{(c)}}
  \end{overpic}
  \begin{overpic}[width=.3\linewidth]{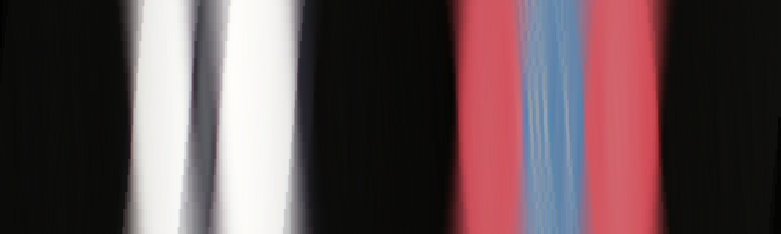}
  \end{overpic}
  \begin{overpic}[width=.3\linewidth]{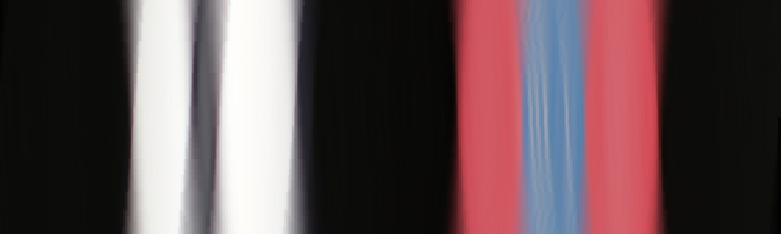}
  \end{overpic}

  \vspace{2pt}
  \begin{overpic}[width=.3\linewidth]{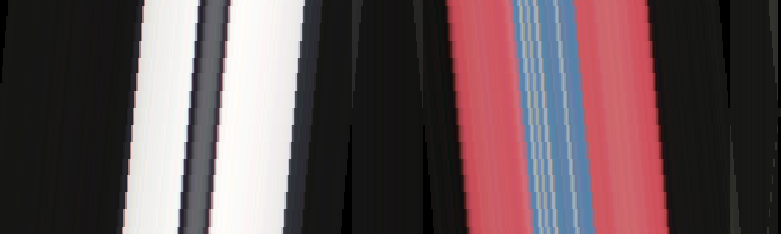}
    \put(-15,0){\rotatebox[origin=tl, units=360]{90}{(d)}}
    \put(40,-15){\color{black}{N=3}}
  \end{overpic}
  \begin{overpic}[width=.3\linewidth]{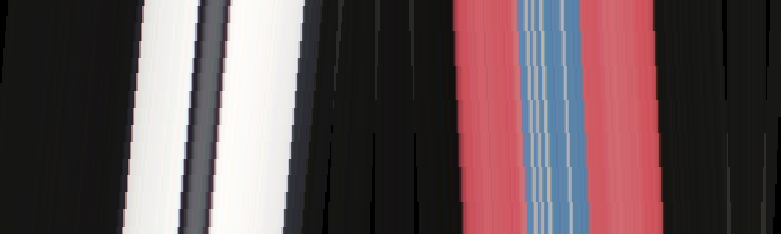}
    \put(40,-15){\color{black}{N=5}}
  \end{overpic}
  \begin{overpic}[width=.3\linewidth]{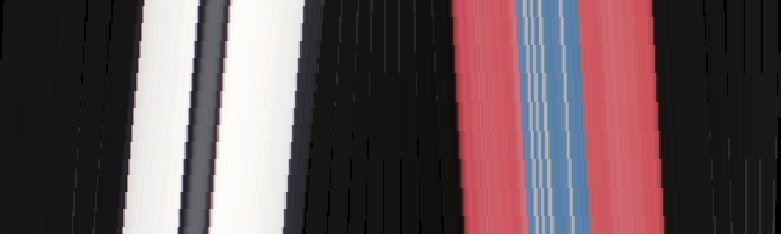}
    \put(40,-15){\color{black}{N=11}}
  \end{overpic}

  \vspace{5pt}
\caption{The reconstructed EPI images of the light field from various number of photos with (a,c) the conventional method and (b,d) the proposed method respectively. (a)(b) represent the light field of $L(x, 0.6y_{max}, \xi, 0)$ and (c)(d) represent $L(x, 0.2y_{max}, \xi, 0)$.}
\label{fig_exp_lf_NI}
\end{figure}
Fig.~\ref{fig_exp_lf_NI} shows the reconstructed EPI images of the light field from various number of photos, where (a)(c) performed with the conventional method and (b)(d) with the proposed method. Fig.~\ref{fig_exp_lf_NI}(a)(b) represent the reconstructed light field of $L(x, 0.6y_{max}, \xi, 0)$ and Fig.~\ref{fig_exp_lf_NI}(c)(d) represent $L(x, 0.2y_{max}, \xi, 0)$. Fig.~\ref{fig_exp_lf_NI}(a)(c) show obvious blur, and the blur behaves too severely while the number of photos used to calculate the light field is large, it even modifies the slope of the EPI images. Conversely, the EPI images of the proposed reconstructed light field are affected slightly by the number of the captured images. The corresponding images reconstructed from the light fields focused at the front object are shown in Fig.~\ref{fig_exp_lf_NI}. Fig.~\ref{fig_exp_lf_NI}(a) and (b) represent the conventional and proposed refocused images respectively. As our expectation, the proposed refocused images perform better quality than the conventional one. This is because the defocus noise in conventional light field is inherited in the refocused images but not in the proposed one. 

\begin{figure}[htb]
  \centering
  \vspace{1pt}
  \begin{overpic}[width=.3\linewidth]{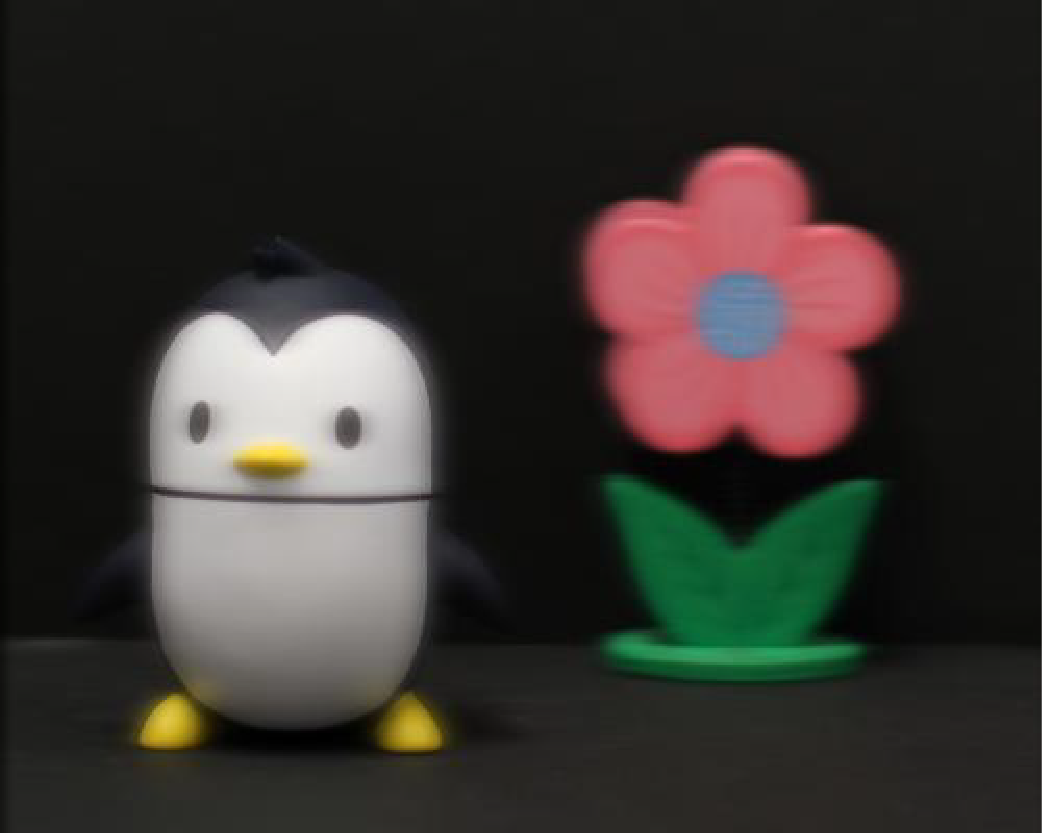}
    \put(-15,20){\rotatebox[origin=tl, units=360]{90}{(a)}}
  \end{overpic}
  \begin{overpic}[width=.3\linewidth]{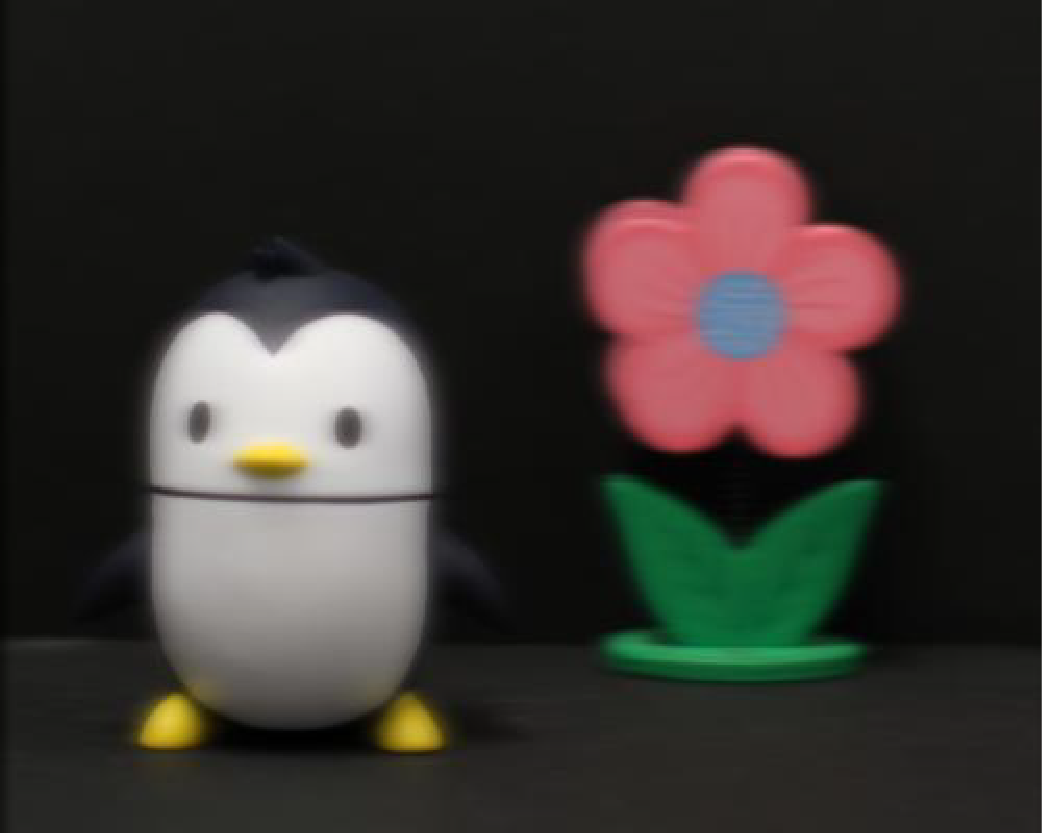}
  \end{overpic}
  \begin{overpic}[width=.3\linewidth]{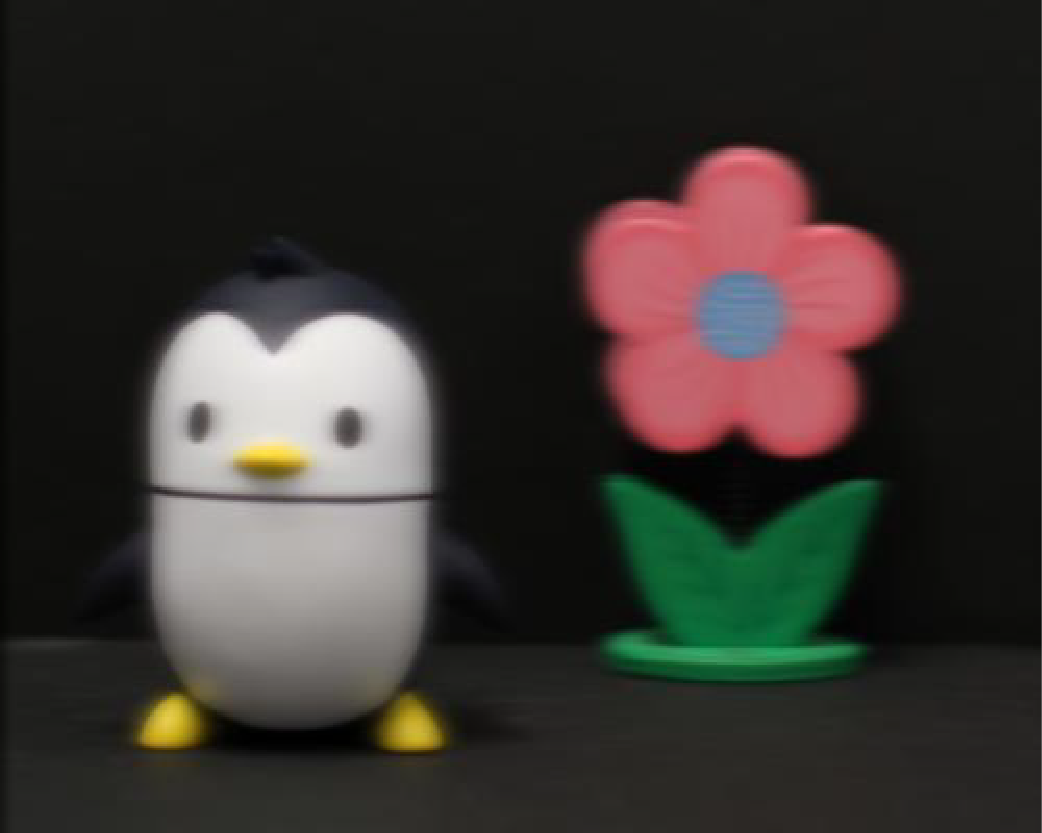}
  \end{overpic}

  \vspace{5pt}
  \begin{overpic}[width=.3\linewidth]{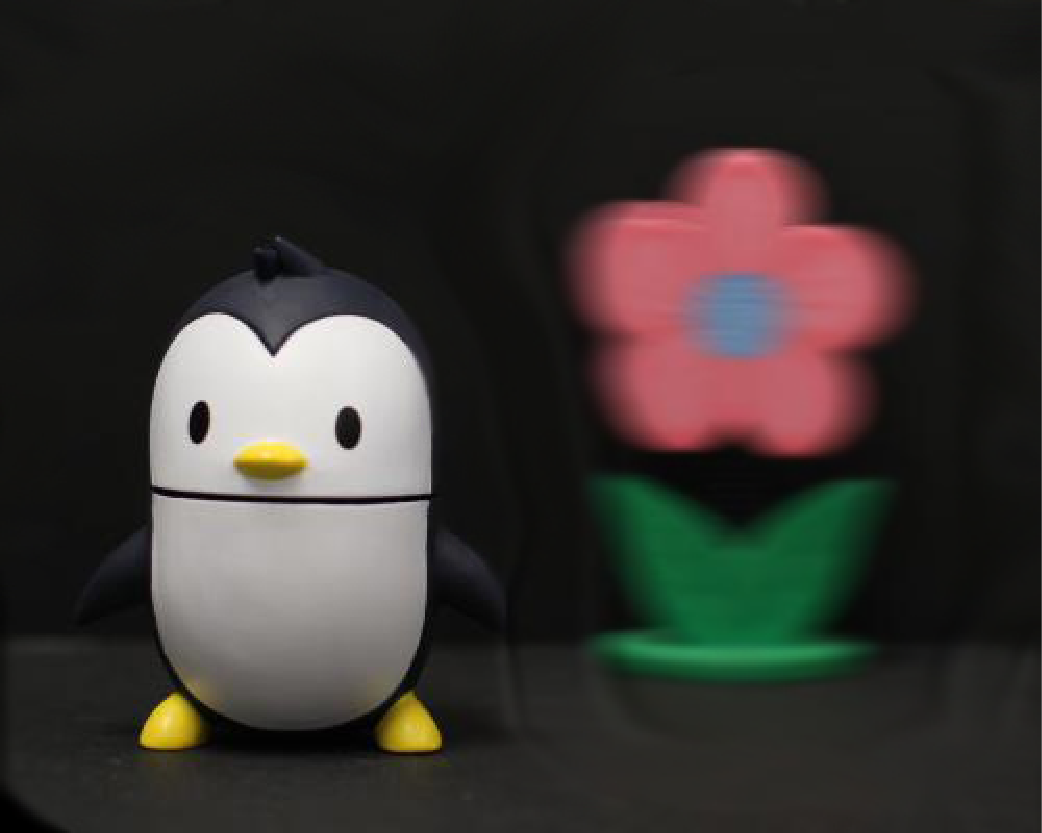}
    \put(-15,20){\rotatebox[origin=tl, units=360]{90}{(b)}}
    \put(40,-15){\color{black}{N=3}}
  \end{overpic}
  \begin{overpic}[width=.3\linewidth]{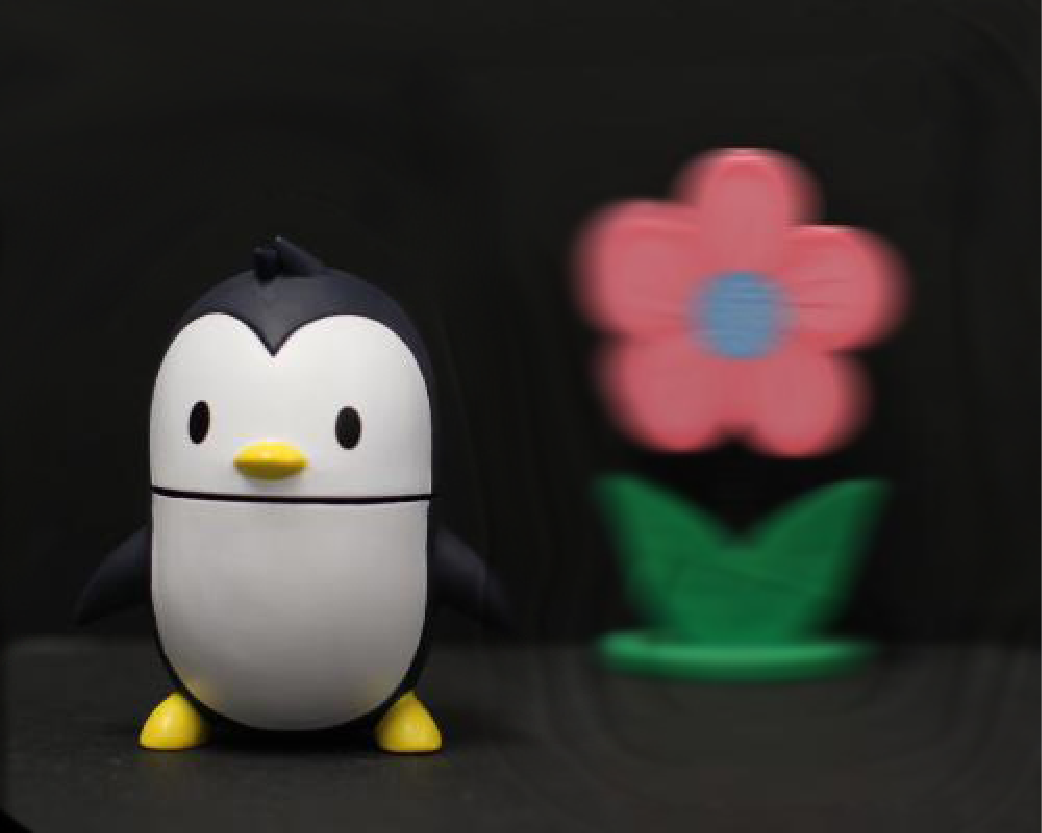}
    \put(40,-15){\color{black}{N=5}}
  \end{overpic}
  \begin{overpic}[width=.3\linewidth]{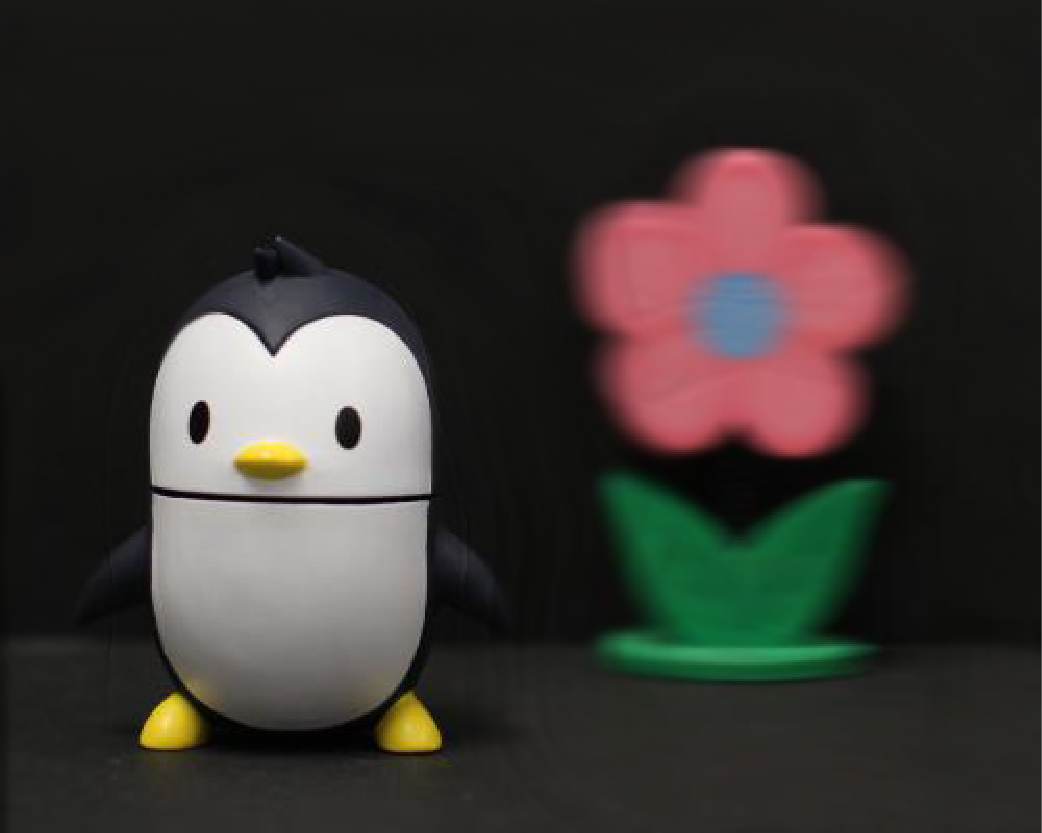}
    \put(40,-15){\color{black}{N=11}}
  \end{overpic}

  \vspace{5pt}
\caption{The refocused images at the front image plane from the calculated light field with (a) the conventional method and (b) the proposed method respectively.}
\label{fig_exp_refocus}
\end{figure}
\section{Conclusions}
We analyzed the noise in the light field reconstruction based on the focal plane sweeping technique. From the analysis, we found that the noise in the reconstructed light field is coming from the accumulation of the defocus noise in the captured photographic images. Therefore it becomes severe with the increasing number of the captured images and the NA of the camera. These are the reasons that limits the application of the light field that calculated with the conventional focal plane sweeping technique. Based on the analysis, we proposed a method to optimize the reconstructed light field by a previous digital deblurring process on the captured photographic images. The proposed method almost eliminates the noise in the reconstructed light field no matter how many captured images we used to calculate it. The simulation and experimental results verified our proposed method.

\section*{Funding Information}
\textbf{Funding.} This work is supported by the National Natural Science Foundation of China~(NSFC)~(Grant Nos. 61377005, 61327902), the Recruitment Program of Global Youth Experts, and the Research Grants Council of the Hong Kong Special Administrative Region, China, under Project N\_HKU714$/$13 through the NSFC/RGC scheme.

\end{document}